\documentclass[letterpaper, dvips, 10 pt, conference]{ieeeconf}  %

\IEEEoverridecommandlockouts                              %

\usepackage{cite}
\usepackage{amsmath,amssymb,amsfonts}
\usepackage{algorithm}  %
\usepackage{algorithmic}
\usepackage{graphicx}
\usepackage{textcomp}
\usepackage{xcolor}
\usepackage{comment}  %
\usepackage[utf8]{inputenc} %
\usepackage[T1]{fontenc}    %
\usepackage{booktabs}
\def\BibTeX{{\rm B\kern-.05em{\sc i\kern-.025em b}\kern-.08em
    T\kern-.1667em\lower.7ex\hbox{E}\kern-.125emX}}
\usepackage[hang,small,bf]{caption}
\usepackage[subrefformat=parens]{subcaption}
\newcommand{\argmax}{\mathop{\rm argmax}\limits}

\renewcommand{\max}{\mathop{\rm max}\limits}
\renewcommand{\min}{\mathop{\rm min}\limits}

\title{\LARGE \bf
Behaviorally Diverse Traffic Simulation \\via Reinforcement Learning}

\begin{document}

\author{Shinya Shiroshita$^{1}$, Shirou Maruyama$^{1}$, Daisuke Nishiyama$^{1}$, Mario Ynocente Castro$^{1}$,\\ Karim Hamzaoui$^{1}$, Guy Rosman$^{2}$, Jonathan DeCastro$^{2}$, Kuan-Hui Lee$^{2}$ and Adrien Gaidon$^{2}$%
\thanks{$^{1}$Preferred Networks, Inc., Japan.,
        {\{shiroshita, maruyama, dnishiyama, marioyc, karim\}@preferred.jp}}
        \thanks{$^{2}$Toyota Research Institute, U.S.,
        {\{guy.rosman, jonathan.decastro, kuan.lee, adrien.gaidon\}@tri.global}}%
}

\maketitle
\thispagestyle{empty}
\pagestyle{empty}

\begin{abstract}
Traffic simulators are important tools in autonomous driving development.
While continuous progress has been made to provide developers more options for modeling various traffic participants, tuning these models to increase their behavioral diversity while maintaining quality is often very challenging.
This paper introduces an easily-tunable policy generation algorithm for autonomous driving agents.
The proposed algorithm balances diversity and driving skills by leveraging the representation and exploration abilities of deep reinforcement learning via a distinct policy set selector.
Moreover, we present an algorithm utilizing intrinsic rewards to widen behavioral differences in the training.
To provide quantitative assessments, we develop two trajectory-based evaluation metrics which measure the differences among policies and behavioral coverage.
We experimentally show the effectiveness of our methods on several challenging intersection scenes.

\end{abstract}

\section{Introducion}

Modeling driving policies is an important and challenging problem in autonomous vehicles research, with a growing range of promising applications.
One such example is the use in behavior prediction modules, which aim to help automated vehicles estimate future trajectories of other vehicles and take necessary safety measures.
Another application of interest is traffic simulation, which we explore in this work.

Modeling human drivers is an important step towards a better understanding of the complex interactions among all traffic participants. 
One difficulty is reconciling driving skills and diversity at the same time. Human drivers are generally skilled, considering the complexity of the driving task. Conversely, they also display a variety of driving characteristics, including aggressive, conservative, or careless types, which make the driving behavior sub-optimal. Therefore, it is important to strike a balance between expanding driving behavioral diversity while maintaining high driving skills.

\begin{figure}[t]
 \centering
 \begin{minipage}[b]{0.65\linewidth}
  \centering
  \includegraphics[keepaspectratio, width=6.0cm]
  {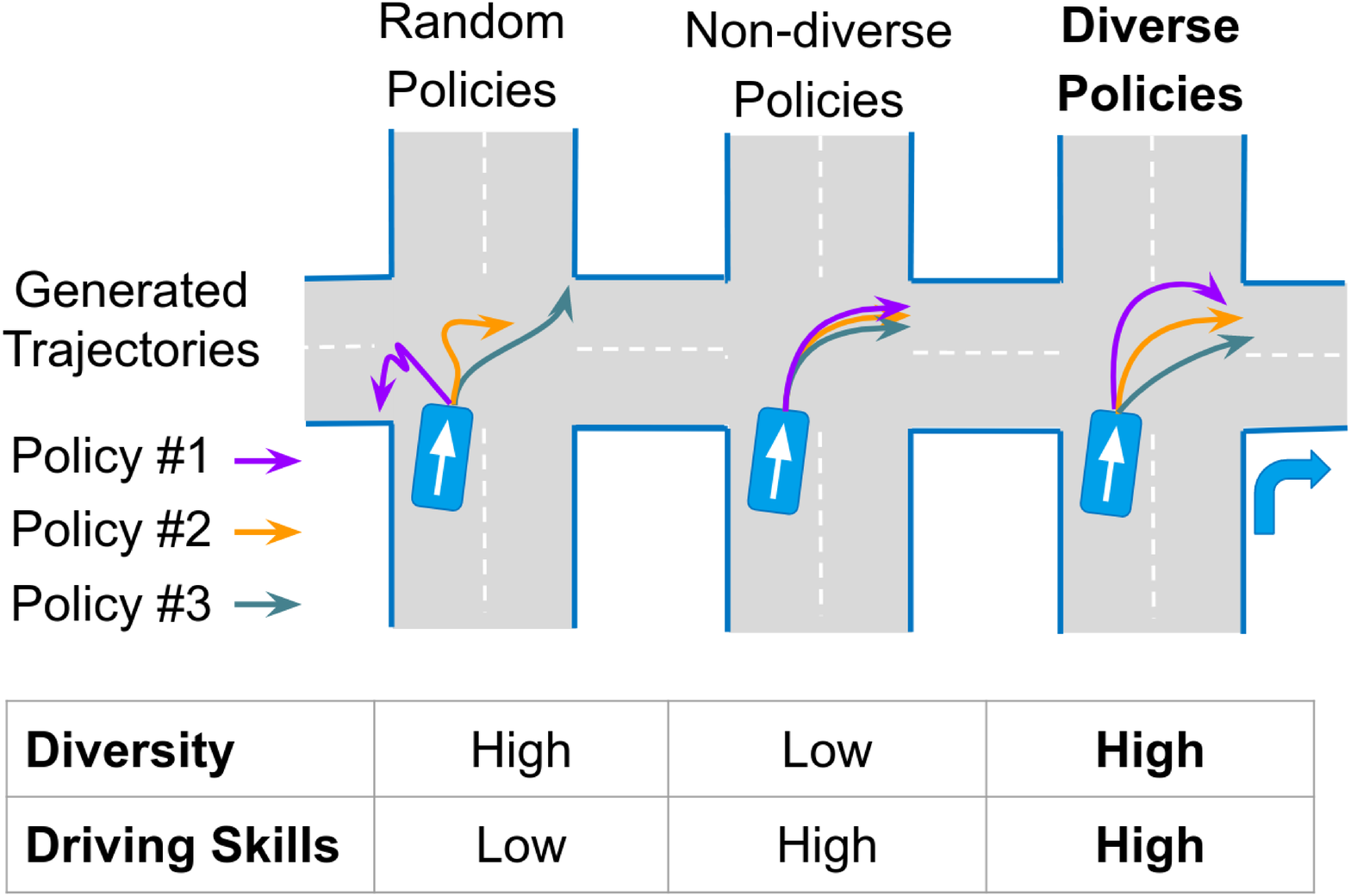}
  \subcaption{}\label{fig:diversity-and-driving-skill}
 \end{minipage}
 \begin{minipage}[b]{0.3\linewidth}
  \centering
  \includegraphics[keepaspectratio, width=3.0cm]
  {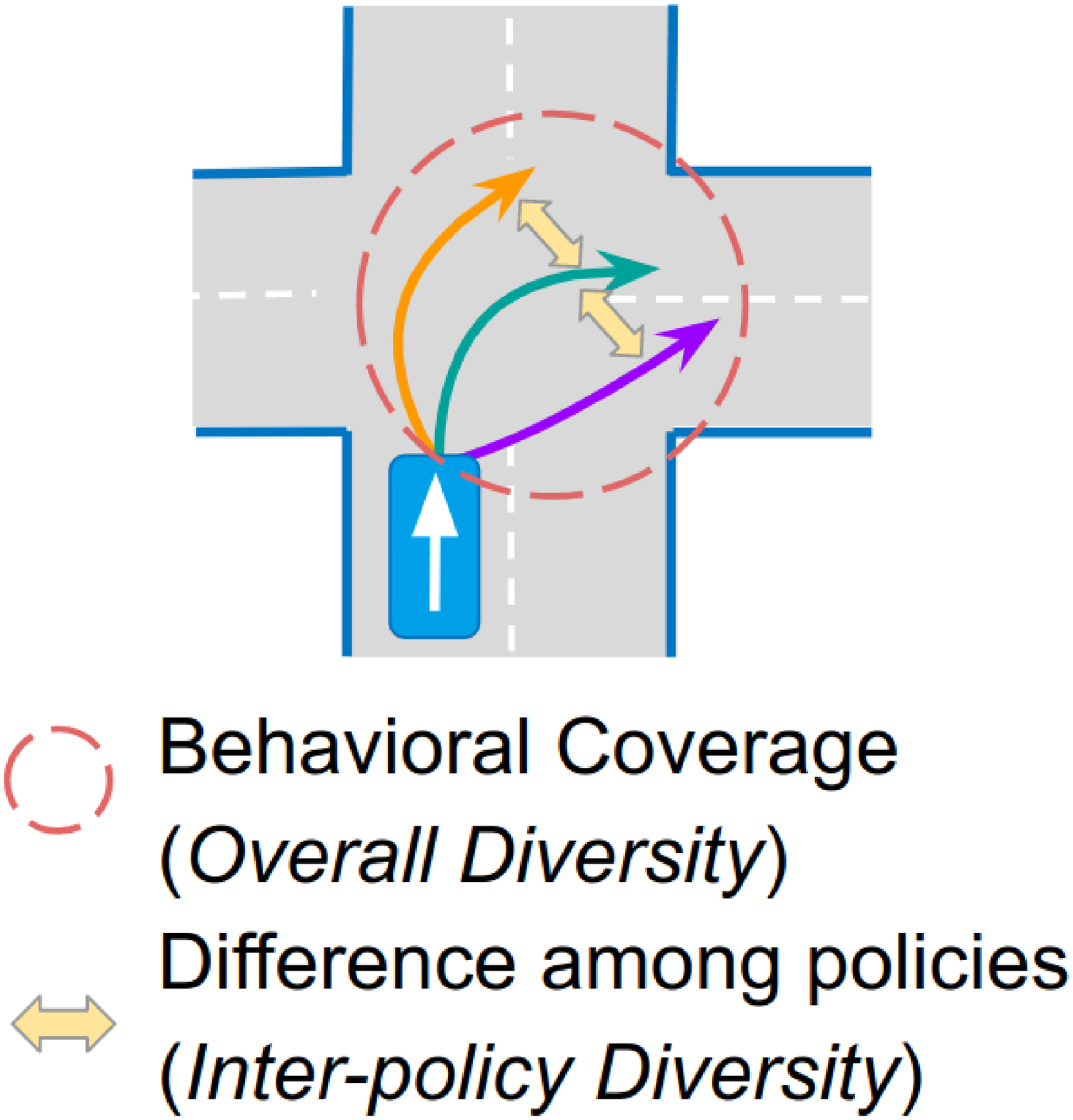}
  \subcaption{}\label{fig:diversity-metrics}
 \end{minipage} 
 \caption{
 \textbf{(a)} Comparison of random policies, non-diverse policies, and diverse policies and their driving skills. \\
 \textbf{(b)} Inter-policy diversity and overall diversity.
 }
 \label{nav-right-turn}
\end{figure}

As shown in Fig.\ref{fig:diversity-and-driving-skill},
one way to impose diversity in a simulator is to use random policies, but the resulting vehicles will have non-sensical trajectories and become useless for simulating realistic traffic situations. Another way is to use a cost function that expresses driving skills when modeling a policy, which will result in the intended driving type.
However, trying to produce various types of driving behaviors using such a method usually results only in minor deviations from optimal behavior.
The goal of our research is to acquire behaviorally diverse driving polices while maintaining high driving skills.

Various approaches have been proposed for diverse behavior modeling.
SUMO~\cite{sumo} provides hand-tuned rule-based policies limited to scripted trajectories in complex scenes such as intersections.
Data-driven approaches (see \cite{social_gan,deo2018multi,salzmann2020trajectron++} and references therein) have difficulty in representing novel behaviors.
Previous reward shaping methods (e.g. Hu~et~al.~\cite{interaction_aware}) made use of reinforcement learning (RL) as a way of generating diversity. However, manually adjusting the reward function to produce various driving policies is difficult for complex traffic scenes such as interactions.  %
In this work, our main contribution is an RL algorithm using intrinsic rewards and a highly expressive model to generate behaviorally diverse driving policies (Fig.~\ref{fig:diversity-metrics}) without requiring additional tuning per target behavior.

To define and reason about the diversity of a set of driving policies, we propose as our second contribution two evaluation metrics based on generated trajectories, reflecting both higher-order decisions and lower-order controls (Fig.~\ref{fig:diversity-metrics}).
Our first metric, Inter-Policy Diversity, measures how different two policies are from each other (Fig.~\ref{fig:diversity-metrics}).
Our second metric, Overall Diversity, represents the overall behavioral coverage (Fig.~\ref{fig:diversity-metrics}).
Since the difference in trajectories largely depends on the traffic scene, including the behavior of other vehicles, both metrics calculate an average of the trajectory distances across multiple traffic scene simulations.

To produce diverse policies with high driving skills, our method first generates various snapshots leveraging the high exploration ability of reinforcement learning, then uses our inter-policy diversity metric to select snapshots where policies express the most distinct behavior.
For the snapshot selection, we filter out those of lower success rates and choose policies by a method similar to Farthest Point Sampling (FPS)~\cite{fps}.
Since diversity from policy selection relies on candidate policies, it is crucial to generate diverse candidates.
To acquire them, we use large scale exploration and a variant of Diversity-Driven Exploration (DDE)~\cite{dde} to enlarge the differences among training policies.

We evaluate our approach on an intersection simulation environment to confirm that it efficiently selects diverse policies.
We also qualitatively check that it produces a variety of cooperation models.

Our contributions can be summarized as follows:
\begin{itemize}
    \item We propose a method of acquiring policies that can balance driving skills and diversity.
    \item We develop trajectory-based diversity evaluation measures for a set of driving policies.
    \item We show the effectiveness of our method for diverse policy acquisition through experiments using a simulation platform which can model vehicle-to-vehicle interactions at intersections.
\end{itemize}
One application of this work is conducted within another work of ours on planner testing~\cite{pfn_planner_test}, where diverse policies are used in surrounding vehicles aiming to detect more diverse failure cases.

The remaining parts are organized as follows.
Section \ref{sec: related_works} and \ref{sec: preliminaries} mention related works and preliminaries, respectively.
Our main contribution is described in the following three sections:
Section \ref{sec: evaluation metrics} for the definition of diversity evaluation metrics,
Section \ref{sec: diversity acquisitions} for the explanation of our policy generation algorithms optimizing the metrics,
and Section \ref{sec: simulation_environments} for the simulator to evaluate our approach's effectiveness.
Section \ref{sec: experiments} summarizes the experimental setting and results.
Section \ref{sec: conclusion} concludes our work.

\section{Related works} \label{sec: related_works}

\subsection{Driving policies in existing traffic simulators}
Many simulators have been developed focusing on various aspects such as
photo-realistic views with a variety of traffic participants (CARLA~\cite{carla}), bird-view traffic flow (SUMO~\cite{sumo}),
light-weighted simulation environment (FLUIDS~\cite{fluids}),
multi-agent framework (CoInCar-Sim~\cite{coincar_sim}),
and benchmark scenarios (CommonRoad~\cite{common_road}).

In terms of driving policies, however, the behavioral diversity of other vehicle policies is not a major focus.
Existing policies consist of simple rule-based policies~\cite{carla, sumo, fluids, coincar_sim, common_road} while some make use of reinforcement learning and imitation learning~\cite{carla}.
However, in all of them, behavioral diversity can be obtained by specifying it manually,
and it is not shown how to select a diverse policy set.
Our research purpose is to establish algorithms that can generate policies for these simulators, which can sustain diversity and good driving skills without having to manually specify one-by-one.

\subsection{Diversity acquisition for driving policies}
Hu~et~al.~\cite{interaction_aware} and Bacchiani~et~al.~\cite{microscpic_multi_agent_rl} generate diverse driving policies by controlling the environmental settings (especially rewards) in merging and roundabout scenes, respectively.
However, these methods depend on specifying such settings manually,  and they do not provide a quantitative comparison of their obtained diversity.
In contrast, our work puts more focus on diverse policy acquisition efficiency and its quantitative evaluation.

In the RL field, diversity appears as a policy exploration tool.
Diversity-Driven Exploration (DDE)~\cite{dde} adds a bonus reward when a policy takes different actions from those taken by previous policies.
Stein Variational Policy Gradient (SVPG)~\cite{svpg} maximizes the entropy of policy parameters to make policies behave differently.
Diversity is All You Need (DIAYN)~\cite{diayn} combines a policy with a discriminator network to distribute states to each action type.
See also Aubret~et~al.~\cite{intrinsic_reward_survey} for a survey about intrinsic rewards.
To the best of our knowledge, our work is the first to investigate RL with intrinsic rewards for traffic simulation tasks. These RL algorithms generally focus on improving benchmark scores instead of providing a diverse simulation platform.

\subsection{Diversity evaluation metrics}
In the field of behavior prediction,
diversity is measured through the prediction error.
Social GAN~\cite{social_gan} and DROGON~\cite{drogon}, for example, evaluated the minimum prediction error among $n$ samples.
However, this metric requires diverse expert trajectories and corresponding maps in the simulator, which is not applicable to any traffic scenes.
Also, the quality of diversity depends on experts and is not directly measured.
Our metrics, on the other hand, evaluate diversity and behavioral coverage separately.
We also show an approximation of diverse trajectory sets to show coverage without experts.

\section{Preliminaries} \label{sec: preliminaries}

Reinforcement learning is a method to learn policies by interacting with an environment.
It assumes the environments as a \textit{Markov decision process (MDP)}.
Let $\mathcal{S}$, $\mathcal{A}$, $p:~\mathcal{S}~\times ~\mathcal{A}~\times~\mathcal{S}~\to~\mathbb{R}$, $r:~\mathcal{S}~\times~\mathcal{A}~\to~\mathbb{R}$
be the state space, the action space, the transition probability, and the reward function, respectively.
Then, the goal is to find a policy $\pi: \mathcal{S} \to \mathcal{A}$ to maximize accumulated rewards with a discount factor $0 < \gamma < 1$:
\begin{equation}
    J = \sum_{t=0}^{\infty} \gamma^t R_{t, \pi}
\end{equation}
where $R_{t, \pi}$ is the expected reward with the policy $\pi$ at the step $t$.
In traffic environments, the observation $o$ of the ego vehicle is usually not equal to the (global) state
since some information such as occluded objects or other vehicle's policy is invisible to the ego vehicle.
Therefore, the policy $\pi$ becomes a mapping from the set of observations to the set of actions.
Section \ref{sec: simulation_environments} shows the details of the observation.

\section{Diversity evaluation metrics} \label{sec: evaluation metrics}

We firstly define objective functions for diversity acquisition algorithms shown in the next section.
For the sake of evaluating the discriminability and the coverage of policies, 
we consider two types of diversity measures: \emph{inter-policy diversity} and \emph{overall diversity}.
The first metric, inter-policy diversity, focuses on the difference between the trajectories generated by pairs of policies.
The second metric, overall diversity, measures the difference between the trajectories generated by a target policy and some particular set of reference trajectories.

Our metrics consider the distances of generated trajectories for each \emph{scenario}, a setting of the surrounding environment, such as a map, initial positions, and other vehicles' policies.
A \emph{trajectory} $\tau = ((x_t, y_t))_{t=1, ..., |\tau|}$ is a sequence of two dimensional coordinates.
We employed the average Euclidean distance for two trajectories $\tau, \tau'$ like
\begin{equation} \label{eq: trajectory distance}
    d(\tau, \tau') = \frac{1}{T} \sum_{t=1}^T ||\tau(t) - \tau'(t)||_2
\end{equation}
where $T = \min\{|\tau|, |\tau'|\}$.
For a scenario $s$ and a policy $\pi$, let $\tau_s(\pi)$ be a trajectory generated by $\pi$ in $s$.
We assume that each scenario also fixes random seeds to ensure deterministic behavior, and thus there is a mapping from the ego policy to the resulted trajectory.

\subsection{Inter-policy diversity}

Let $S, \Pi$ be a set of scenarios and a policy set, respectively.
Let $S_\pi \subseteq S$ be a set of scenarios where a policy $\pi$ succeeds the mission to reach a given goal within a time limit without collision (see Section \ref{subsec: experimental setting} for more details), %
and similarly let $\Pi_s \subseteq \Pi$.
Note that for any scenario $s$ and any policy pair $\pi, \pi'$,
we assume that $\Pi_s \neq \emptyset$ and $S_\pi \bigcap S_{\pi'} \neq \emptyset$.
Exceptions rarely happen since polices with low success ratios are removed at the policy selection step.

We define the inter-policy diversity $D_{IP}$ of a policy set $\Pi$ by the average distance of two policy pairs like
\begin{equation}
    D_{IP}(\Pi) = \frac{1}{|\Pi| (|\Pi| - 1)} \sum_{\pi \in \Pi}\sum_{\pi' \in \Pi\setminus\{\pi\}} D_{IP}(\pi, \pi')
\end{equation}
where
\begin{equation}
    D_{IP}(\pi, \pi') = \frac{1}{|S_\pi \bigcap S_{\pi'}|} \sum_{s \in S_\pi \bigcap S_{\pi'}} d(\tau_{s}(\pi), \tau_s(\pi'))
\end{equation}
means an average trajectory distance of two policies $\pi, \pi'$.
Our work utilizes the diversity for discriminability evaluation and a basis for the policy set selection.

\subsection{Overall diversity}\label{subsec: overall_diversity}

Overall diversity calculates the distance between the obtained trajectory and the expected trajectories called \emph{reference trajectories} $\mathcal{T}$.
To capture both the density and the spatial similarity of trajectories, we employ the Wasserstein-1 distance as a distance measure of distributions.
For each scenario $s$, let $\tau_s(\Pi) = \{\tau_s(\pi) | \pi \in \Pi_s\}$ be a set of successful trajectories of $\Pi$ in $s$.
Then, we define the overall diversity $D_{OA}^{\mathcal{T}, s}$ of a policy set $\Pi$ as
\begin{equation}
    D_{OA}^{\mathcal{T}, s}(\Pi) = \inf_{\gamma \in \Gamma(\tau_s(\Pi), \mathcal{T})} \mathbb{E}_{(\tau, \tau') \thicksim \gamma}[d(\tau, \tau')]
\end{equation}
where $\Gamma(\tau_s(\Pi), \mathcal{T})$ is a set of all possible joint distributions of $\tau_s(\Pi)$ and $\mathcal{T}$.
We finally obtain the overall diversity $D_{OA}^{\mathcal{T}}$ of the policy set by the average over scenarios like
\begin{equation}
    D_{OA}^{\mathcal{T}}(\Pi) = \frac{1}{|S|} \sum_{s \in S} D_{OA}^{\mathcal{T}, s}(\Pi_s).
\end{equation}
Note that a lower value indicates better performance.

Since counting all possible trajectories is infeasible, we need to set a well-approximated $\mathcal{T}$ covering behaviors as diverse as possible.
We modeled $\mathcal{T}$ as a set of \emph{Brownian bridges} modified to take into consideration vehicle dynamics and surrounding vehicles.
The process of generating the bridges consists of the following steps.
\begin{enumerate}
    \item Define an expected right-turn trajectory manually. A generator draws diverse trajectories along with that trajectory.
    \item Draw \emph{target trajectories}. These trajectories are created by perturbating the expected trajectory with longitudinal and lateral movements, respectively.
    The longitudinal movement is for speed variation and the lateral movement is for steering variation, respectively.
    Each movement is generated by Brownian bridges, which are stochastic processes whose initial and end points are fixed, to restrict the destination and arrival time.
    \item Run a Proportional control (P-control) vehicle trying to mimic the target trajectories. Let $\tau^\text{target}$ and $\tau$ denote the target and the converted trajectory, respectively. The vehicle at time $t$ will try to move from point $\tau(t)$, in $\nu$ seconds, to the point $\tau^\text{target}(t+\nu)$ which is the point in the original trajectory $\nu$ seconds later.
    
\end{enumerate}
We run the above procedure in the intersection environment shown in Section \ref{sec: simulation_environments}.
If the agent reaches the goal without collision, the trace of the vehicle's center is added to the output.
Fig. \ref{fig:reference_trajectory} in the experimental section shows an example of the generated set of trajectories in one scenario.

\section{Diversity Acquisition Algorithms} \label{sec: diversity acquisitions}

This section shows the diversity acquisition modules that generate a set of policies with better inter-policy and overall diversity values
defined in the previous section.
Since there exists a trade-off between behavioral diversity and driving quality,
we put a restriction on policies where their mission success rates must not be worse than a threshold~$\delta$.

Our strategy is as follows:
we train a set of $n (\geq 2)$ policies $\Pi$ simultaneously
and select snapshots (not only the latest ones but also those during training) in order to maximize the inter-policy diversity.
Details are explained in Section \ref{subsec: policy selection}.
In the case of environments where random policies tend to converge into the same policy, we also propose an algorithm which uses intrinsic rewards in Section \ref{subsec: dde algorithm}.

\subsection{Diverse policy selection} \label{subsec: policy selection}
Reinforcement learning produces a variety of policies through the exploration process guided  %
by initial parameters and randomness during training.
The basic strategy is to first create many snapshots of the policies during their training process.
Let $\Pi_{all}$ be a set of candidate policies that consists of snapshots generated during the training process.
Firstly, we remove policies from $\Pi_{all}$ with a driving score (e.g. success ratio) is less than the threshold $\delta$ to filter out the policies with low driving skills.  
Let $\Pi \subseteq \Pi_{all}$ be the filtered set, and $k$ be the number of agents to be selected.
Then, our target is to find a subset $Q_k^* \in \mathcal{P}_k(\Pi) = \{Q \mid Q \in 2^{\Pi} \text{ and } |Q|=k\}$ such that %
\begin{equation}
Q_k^* = \argmax_{Q \in \mathcal{P}_k(\Pi)}({D_{IP}(Q)})
\end{equation}
which maximizes the inter-policy diversity.

However, finding an optimal subset from all possible choices is computationally infeasible since $|\mathcal{P}_k(\Pi)|$ exponentially increases according to $k$.
Therefore, we propose a greedy method based on farthest point sampling~\cite{fps} to find an approximation of $Q_k^*$.

Our method first selects one policy at random and repeatedly adds a new one where the average inter-policy diversity with the selected policies is the highest among the remainder.
Algorithm \ref{alg: diverse policy selection} shows the precise procedure.
Although this algorithm considers only inter-policy diversity,
our experimental results suggest that the generated set of policies is also diverse in terms of overall diversity due to the spatially separated selection process.

\begin{algorithm}[t]
    \caption{Diverse policy selection}
    \label{alg: diverse policy selection}
    \begin{algorithmic}
        \REQUIRE{Policies $\Pi_{all}$, scenarios $S$, driving score threshold $\delta$, number of selected policies $k$}
        \ENSURE{Set of $k$ policies $Q_k$}
        \STATE{$\Pi \leftarrow \{\pi \in \Pi_{all} \mid \pi\text{'s driving score is not less than } \delta \}$}
        \STATE{$Q_1 \leftarrow \{\pi_1\}$ where $\pi_1$ is randomly selected from $\Pi$}
        \FOR{$i = 2, ..., k$}
            \STATE{$\pi_i \leftarrow \argmax_{\pi \in \Pi \setminus Q_{i-1}}{\min_{\pi' \in Q_{i-1}}D_{IP}(\pi, \pi')}$}
            \STATE{$Q_i \leftarrow Q_{i - 1} \cup \{\pi_i\}$}
        \ENDFOR
        \RETURN{$Q_k$}
    \end{algorithmic}
\end{algorithm}

\subsection{Diverse policy training with intrinsic rewards} \label{subsec: dde algorithm}

Another way of enhancing diversity is by utilizing intrinsic rewards on different actions to enlarge policy differences, especially for reward settings where policies with different initial parameters are likely to converge into similar ones.
Our approach is a modification of DDE~\cite{dde}.
To put a repulsive force among training policies,
our method runs $m$ training threads in parallel and set intrinsic rewards as
\begin{equation}
    r^\text{(env)} + \mathbb{E}_{\pi' \in \hat{\Pi}\setminus\{\pi\}}[\alpha d_{KL}(\pi, \pi')]
\end{equation}
where $r^\text{(env)}$ is the environmental reward, $\pi$ is a current policy, $\hat{\Pi}$ is the current policy set beyond all threads (while the original DDE used a set of previous policies),
$\alpha$ is the weight of the intrinsic reward, and $d_{KL}(\pi, \pi')$ is the KL divergence of action distributions of $\pi$ and $\pi'$, respectively.

Although the original DDE~\cite{dde} used a tunable $\alpha$, we fixed $\alpha$ in the experiments.

\section{Traffic Simulation} \label{sec: simulation_environments}
To verify our proposed evaluation and method, we have implemented a simulation environment with intersection-based traffic scenes.
This simulator can perform simulation of multiple agents.
The goals of each agent include:
(i) moving along with the given navigation route as fast as possible, 
and (ii) performing collision-free movements to walls and other vehicles.

\subsection{Map}
\begin{figure}
    \centering
    \includegraphics[keepaspectratio, width=5.0cm]
    {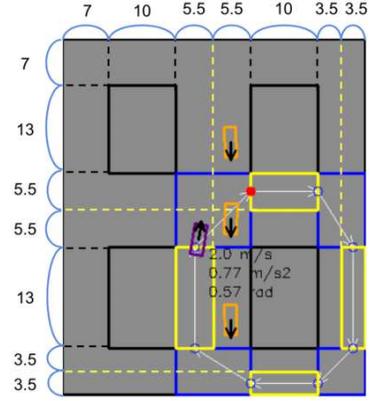}
    \caption{The scales of the simulation map.
    All units are meters.
    The width and length of each vehicle are 1.8 and 4.5 meters, respectively. } \label{fig:map2}
\end{figure}

The map we prepared follows the Japanese traffic rule, 
that is, vehicles on the road drive on the left side of the driving lanes.
Fig.~\ref{fig:map2} shows the scales of the map.
The purple rectangle and orange ones represent the ego vehicle and the other vehicles, respectively.
The black arrow on a vehicle represents its heading.
Each vehicle is given a predefined \emph{navigation route}
consisting of a set of \emph{target arrows} (white arrows) the vehicle is expected to follow.
Each target arrow is combined with a rectangular \emph{zone} to define rewards for the vehicle.
The zone has two types: straight-line zones (yellow rectangles) and intersection zones (blue rectangles) with the different reward calculation formula.
The map also contains \emph{walls} (black lines) which vehicles cannot pass beyond.

\begin{figure}
    \centering
    \includegraphics[keepaspectratio, width=7.0cm]
    {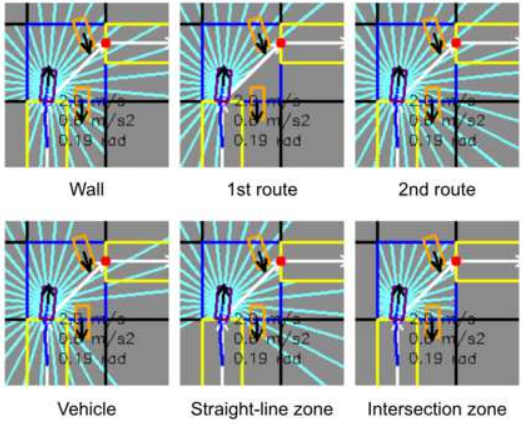}
    \caption{Six types of distances for observation. Each light blue line emitted from the purple rectangle represents the distance of the captioned parameter in the corresponding direction.} \label{fig:lidar2}
\end{figure}

\subsection{Episode termination conditions}
In the evaluation phase, the episode is stopped if a collision occurs, if the ego vehicle reaches a predefined goal area, or if the episode time limit expires.
In training, there is no fixed goal, and the trainee tries to go around the navigation route forever
unless the time is over or it has more than 150 collisions in an episode.

\subsection{Vehicle Dynamics}
The simulator regards each vehicle as a rectangle running under the kinematic bicycle model~\cite{bicycle_model}.
The length from the gravity center to each wheel is equal to the half of the vehicle length.
The control input for each vehicle at time $t$ is specified as $u_t = (\phi_t, a_t)$, where $\phi_t \in [-0.785, 0.785]$ is the steering angle ($\mathrm{rad}$) and $a_t \in [-1.0, 1.0]$ is the acceleration ($\mathrm{m/s^2}$), respectively.
According to the control input, the vehicle state transition from $c_t = (x_t, y_t, \theta_t, v_t)$ to $c_{t + \varDelta t}$ is computed
where $x_t$ and $y_t$ represent the two dimensional coordinates of the vehicle center ($\mathrm{m}$),
$\theta_t$ is the heading of the vehicle ($\mathrm{rad}$), $v_t$ is the velocity of the vehicle ($\mathrm{m/s})$, and $\varDelta t$ is time difference between two adjacent frames.
The velocity range is limited to $v_t \in [0.0, 2.0]$, which is sufficient for driving at an intersection.

\subsection{Observation and Action Space}
In terms of observation, the agent acquires six types of distances for each of 32 directions $\mu_{w}, \mu_{r1}, \mu_{r2}, \mu_{v}, \mu_{z^s}, \mu_{z^i} \in\mathbb{R}^{32}$.
Each value, visualized in Fig. \ref{fig:lidar2}, is defined as, the shortest distance (50 meters is maximum) to walls, the closest navigation route, the second closest one, other vehicles, the straight-line zone boundaries, and the intersection zone ones, respectively.
Combining with the last three ego vehicle states $h_{t-\varDelta t}, ..., h_{t-3\varDelta t}$ where $h_t~=~(v_t, a_t, \phi_t)~\in~\mathbb{R}^{3}$,
the overall observation becomes
\begin{eqnarray}
o_t & = &[\mu_{w}, \mu_{r1},\mu_{r2},\mu_{v}, \mu_{z^s}, \mu_{z^i}, \nonumber \\
& \phantom{=} & \phantom{[]} h_{t-\varDelta t}, h_{t-2\varDelta t}, h_{t-3\varDelta t}] \in\mathbb{R}^{201}.
\end{eqnarray}

Actions, on the other hand, consist of nine discrete %
pairs of $(\gamma_1, \gamma_2)$ shown in Table \ref{tbl:correspondence-of-action-types}.
The control input at time $t$ is determined as $u_{t}=(\gamma_1\varDelta t+\phi_{t-\Delta t}, \gamma_2\varDelta t + a_{t-\Delta t}).$

\begin{table}[tb]
\vspace{2mm}
\caption{Correspondence of the action types.}
\centering
\begin{tabular}{|lrr||lrr|} \hline
Action & $\gamma_1$ & $\gamma_2$ & 
Action & $\gamma_1$ & $\gamma_2$ \\\hline
Forward & 0 & 2.5 & Right-forward & 0.628 & 2.5\\
Backward & 0 & -2.5 & Left-forward & -0.628 & 2.5 \\
Right & 0.628 & 0 & Right-backward & 0.628 & -2.5\\
Left & -0.628 & 0 & Left-backward & -0.628 & -2.5\\
Holding & 0 & 0 & & & \\\hline
\end{tabular}
\label{tbl:correspondence-of-action-types}
\vspace{-2mm}
\end{table}

\subsection{Reward Function}
\begin{figure}
    \vspace{2mm}
    \centering
    \includegraphics[width=7cm]{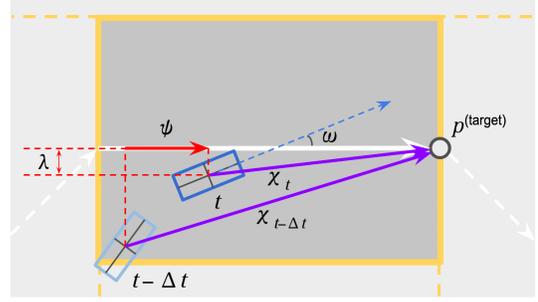}
    \caption{
        Variables used for reward calculation.
        The yellow rectangle represents a zone
        and the dark blue one represents the ego vehicle at time $t$.}
    \label{fig:reward}
    \vspace{-2mm}
\end{figure}

The environmental reward $r^\text{(env)}$ consists of four types of rewards $r^\text{(move)}, r^\text{(collision)}, r^\text{(angle)}, r^\text{(center)}$ as
\begin{equation}
    r^\text{(env)} = r^\text{(move)} + r^\text{(collision)} + r^\text{(angle)} + r^\text{(center)}.
\end{equation}
These rewards are parametrized by non-negative weight coefficients $w^\text{(move)}, w^\text{(collision)}, w^\text{(angle)},\ \text{and}\ w^\text{(center)}$.
Fig. \ref{fig:reward} summarizes variables used for each reward's description.

\subsubsection{Movement Reward}
        $r^\text{(move)}$ is specified based on the progress toward the desired direction, which depends on the zone the ego vehicle is in.
        In the intersection zones,
            \begin{equation}
                r^\text{(move)} = w^\text{(move)} \cdot \max{\{0, \chi_{t-\varDelta t} - \chi_t\}}
            \end{equation}
        where $\chi_{t-\varDelta t}, \chi_{t}$ are the Euclidean distances from the end point $p^\text{(target)}$ of the target arrow to $(x_{t - \varDelta t}, y_{t - \varDelta t}), (x_t, y_t)$, respectively.
        In the straight-line zones,
            \begin{equation}
                r^\text{(move)} = w^\text{(move)} \cdot \psi
            \end{equation}
        where $\psi$ is the travel distance from frame $t - \varDelta t$ to $t$ projected on the target arrow
        ($\psi$ can be negative if the ego vehicle recedes).
        If the ego vehicle exists in neither, $r^\text{(move)}=0$.

\subsubsection{Collision Reward}
        $r^\text{(collision)}$ is defined as
            \begin{equation}
                r^\text{(collision)} =
                \begin{cases}
                    -w^\text{(collision)} & \text{if a collision occurs} \\
                    0 & \text{otherwise}.
                \end{cases}
            \end{equation}

\subsubsection{Angle Reward}
        $r^\text{(angle)}$ is based on the difference between angles of the ego vehicle and the expected route.
        When the agent is in a straight-line zone and $r^\text{(move)} \geq 0$,
        \begin{equation}
            r^\text{(angle)} =  w^\text{(angle)}\cdot(0.5 - (\omega / \pi)^2).
        \end{equation}
        where $\omega$ is the difference ($\mathrm{rad}$) of $\phi_t$ and the target arrow's angle.
        $r^\text{(angle)} = 0$ otherwise.

\subsubsection{Center-line Reward}
        $r^\text{(center)}$ represents how close the ego vehicle is from the route.
        When the agent is in a straight-line zone and $r^\text{(move)} \geq 0$,
        \begin{equation}
            r^\text{(center)} = w^\text{(center)}\cdot(5\cdot \exp(-8 \cdot \lambda^2)- 0.5)
        \end{equation}
        where $\lambda$ is the shortest distance from $(x_t, y_t)$ to the target arrow.
        $r^\text{(center)} = 0$ otherwise.

\subsection{Reinforcement Learning and Neural Network}
\begin{figure}
 \centering
 \begin{minipage}[b]{0.48\linewidth}
  \centering
  \includegraphics[keepaspectratio, width=3.5cm]
  {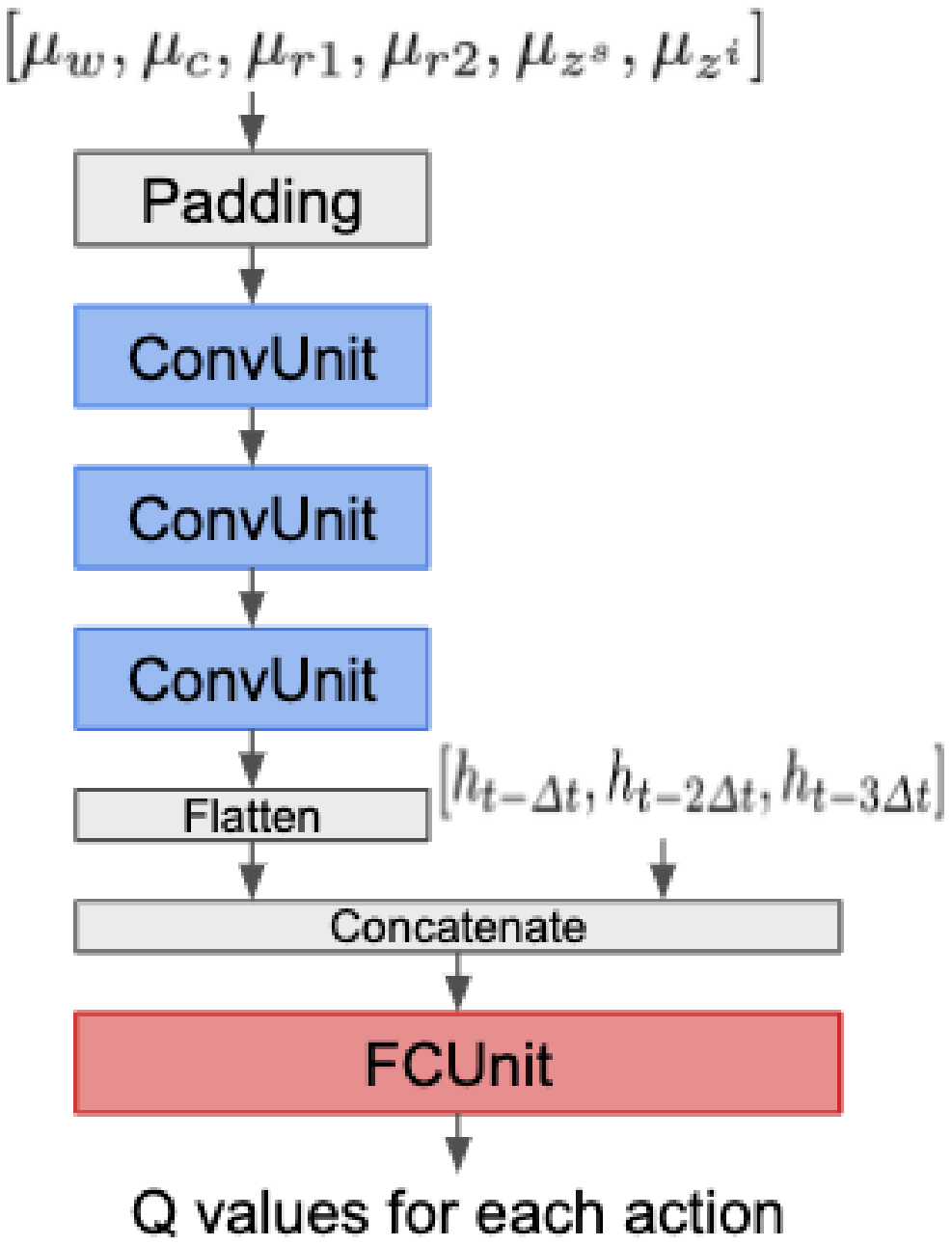}
  \subcaption{}
 \end{minipage}
 \begin{minipage}[b]{0.22\linewidth}
  \centering
  \includegraphics[keepaspectratio, width=1.8cm]
  {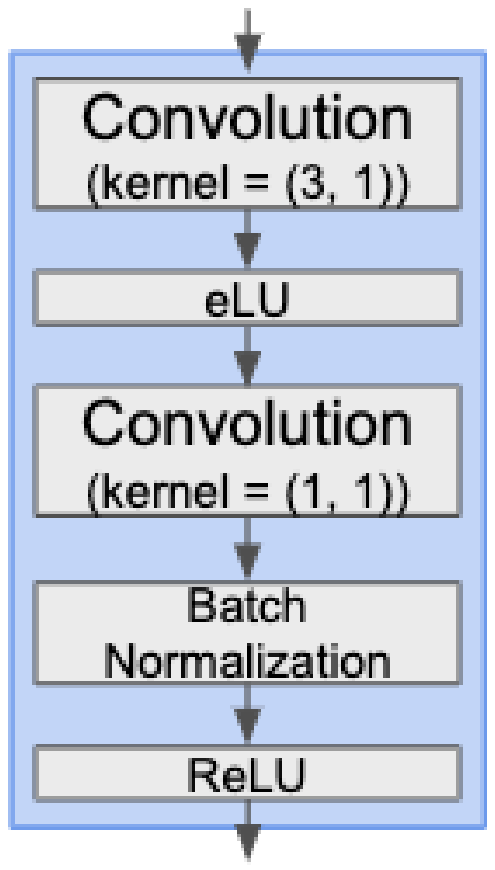}
  \subcaption{}
 \end{minipage}
 \begin{minipage}[b]{0.22\linewidth}
  \centering
  \includegraphics[keepaspectratio, width=1.8cm]
  {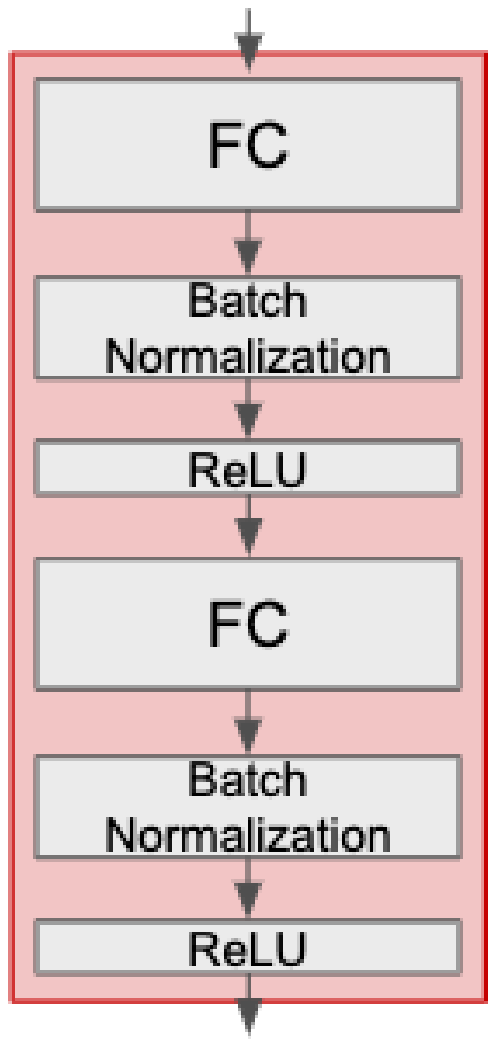}
  \subcaption{}
 \end{minipage}
  \caption{
      \textbf{(a)} The overall architecture.
      The padding layer appends the first few repeats inputs to distances.
      \textbf{(b)} The \emph{ConvUnit}.
      Any convolution layer has 50 hidden channels and strides $(1, 1)$.
      \textbf{(c)} The \emph{FCUnit}.
      Each fully-connected~(FC) layer has 600 hidden channels.
  }
 \label{fig:nn_architecture}
\end{figure}

We used the double PAL as the reinforcement learning algorithm and the distances convolutional network model proposed in the previous work~\cite{pfn_sim}.
The differences are the number of input/output dimensions of the network and the absence of the time series of distance inputs.
The details of the architecture are shown in Fig. \ref{fig:nn_architecture}.
We also use the Adam optimizer~\cite{adam} and the epsilon-greedy method for exploration.

\section{Experiments} \label{sec: experiments}

In order to check the effectiveness of our method, we conducted experiments in both single-agent and multi-agent settings.
Firstly, we show the common experimental settings for both cases.
Then, we show some quantitative results for the diversity acquisition in the single-agent setting,
where the surrounding vehicles ignore the ego vehicle.
Finally, we show qualitative results to confirm that the generated policies can perform various kinds of interactions among multiple agents.

Although the following subsections discuss the right-turn traffic scene (Fig. \ref{nav-right-turn}), we also ran experiments on the straight-straight intersection scene and the 4-directional intersection scene.
The results can be found in the supplemental video.

\subsection{Experimental Setting} \label{subsec: experimental setting}

\begin{figure}
 \vspace{2mm}
 \centering
 \begin{minipage}[b]{0.45\linewidth}
  \centering
  \includegraphics[keepaspectratio, width=2.7cm]
  {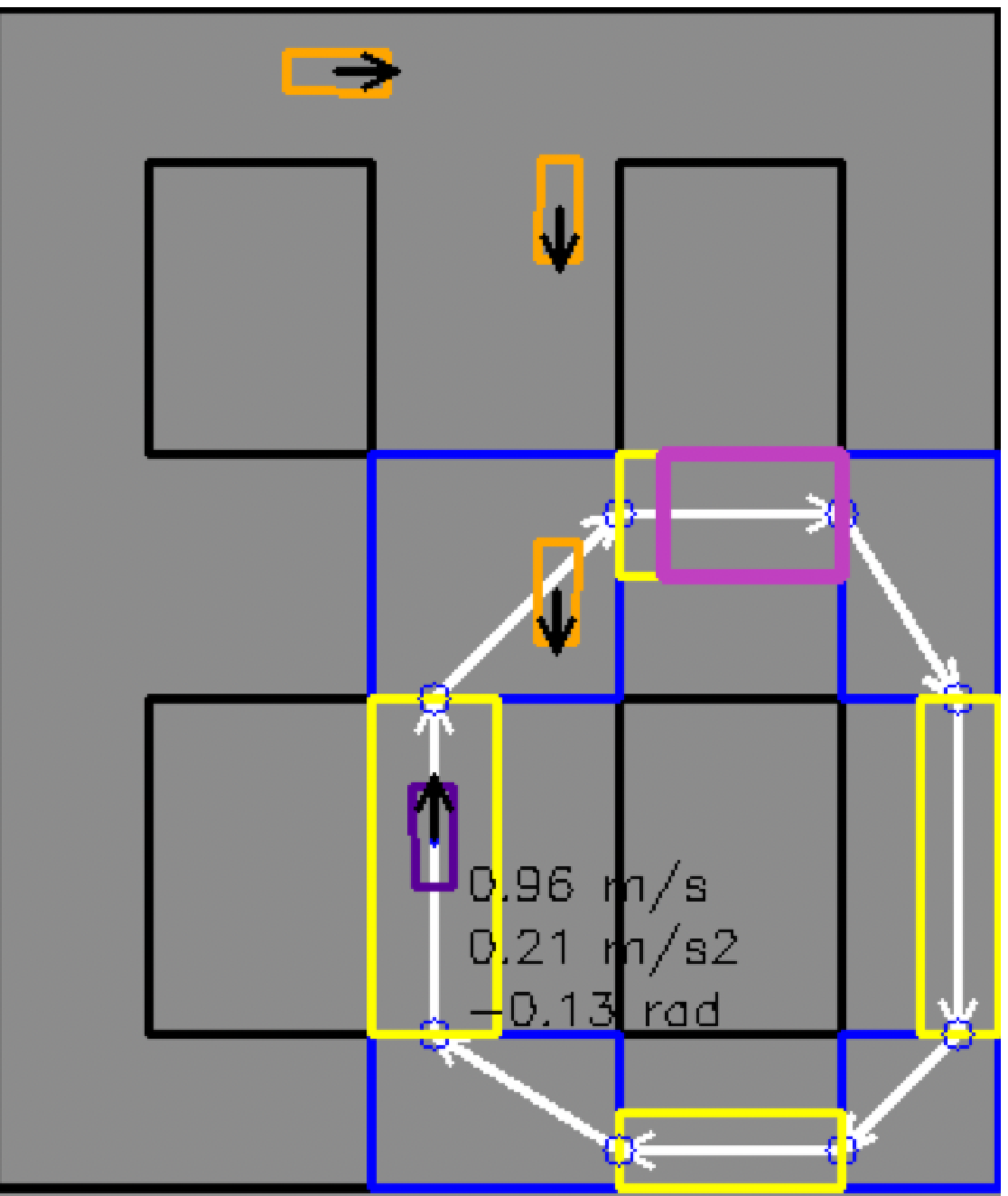}
  \subcaption{Route for the ego vehicle.}\label{nav-for-blue}
 \end{minipage}
 \begin{minipage}[b]{0.45\linewidth}
  \centering
  \includegraphics[keepaspectratio, width=2.7cm]
  {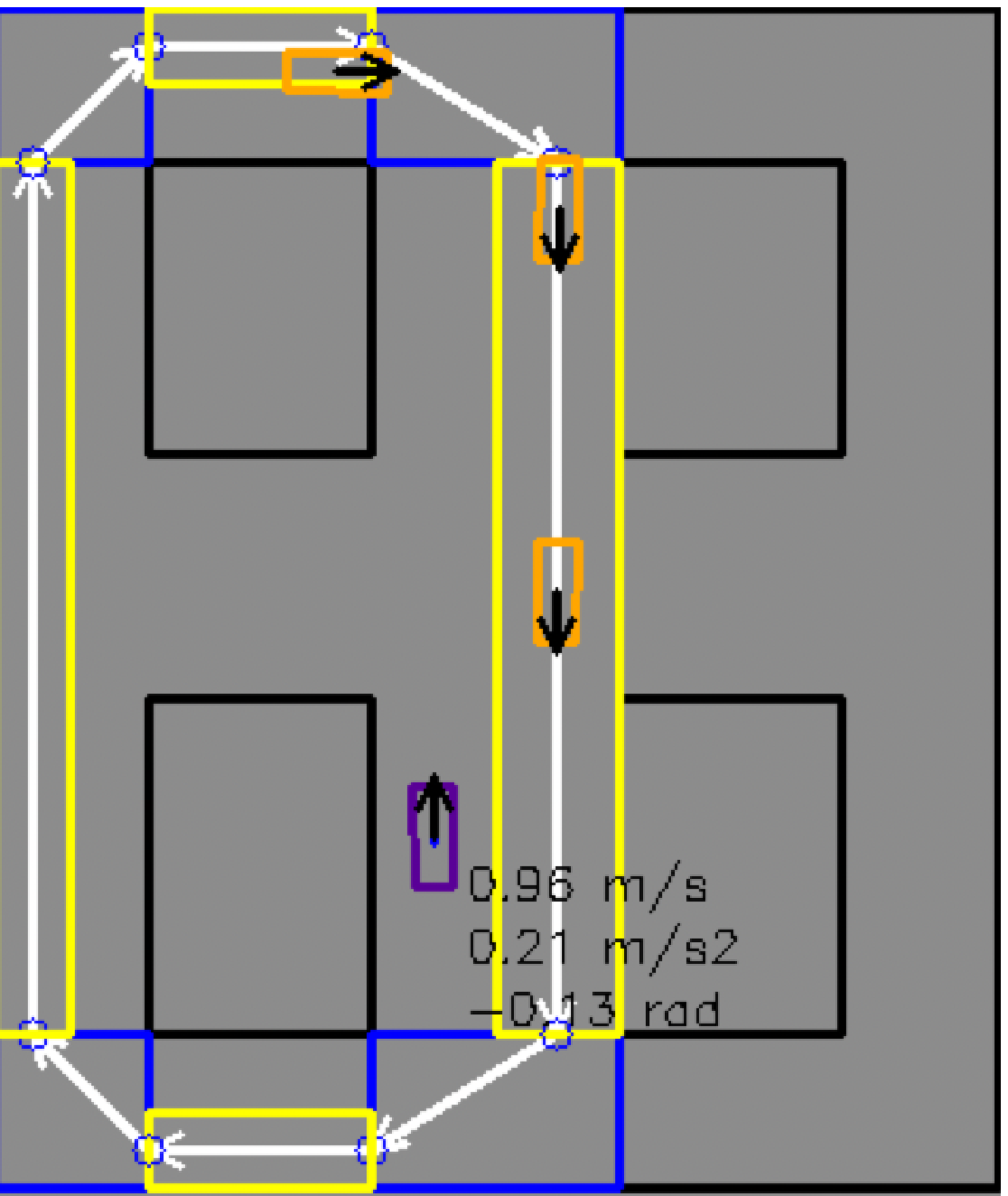}
  \subcaption{Route for the other vehicles.}
 \end{minipage} 
 \caption{An example of initial positions of vehicles and their navigation routes for the right-turn traffic. %
 }\label{nav-right-turn}
 \vspace{-2mm}
\end{figure}

Details of our experiments settings are listed in this subsection.
Fig.~\ref{nav-right-turn} shows the navigation route of the target evaluated vehicle making a right turn (dark purple rectangle) and that of the other vehicles going straight down the opposite lane (orange rectangles).
The mission of the ego vehicle is to reach to the goal area (purple rectangle in Fig. \ref{nav-for-blue}) within 25 seconds without colliding with vehicles nor into walls.
We set ${\varDelta}t~=~0.1$ (thus, the maximum episode length is 250).
The driving score of a policy $\pi$ is defined as the proportion of successful scenarios $S_\pi$, where the agent successfully reaches its goal, to all scenarios $S$. %
In all settings, we set the driving score threshold as $\delta = 90 \%$.

Weight coefficients and hyperparameters are as follows:
\begin{itemize}
    \item $w^\text{(move)}=100$ in all experiments.
    \item $(w^\text{(angle)},~w^\text{(center)}) = (15,~5)$ in the multi-agent settings and $(0,~0)$ otherwise.
    \item $w^\text{(collision)}$ was linearly increased from $0$ to $300$ during the first $300,000$ steps in all settings as in Miyashita~et~al.~\cite{pfn_sim}.
    \item $lr = 0.001$ in learning rate of Adam.
    \item $\epsilon$ of epsilon-greedy method was linearly decreased from $1.0$ to $0.1$ during the first $100,000$ steps.
    \item $\alpha~=~0.01$ which is the coefficient in DDE.
\end{itemize}

During training, the initial position of the cars is perturbated by a random number for each episode.
In order to compare all methods under the same scenarios, we prepared 50 scenarios for evaluation, in which random numbers for the initial positions are fixed.
To generate the candidates, we have run several  training sessions in parallel.
Each session trains an agent to create 150 snapshots of every 20,000 steps out of 3 million training steps.
Consequently, the number of sessions for each candidate set is equal to the number of candidates divided by 150.

We did all the implementation in Python 3. We also used the implementation of Chainer~\cite{chainer} and ChainerRL~\cite{chainerrl} for the RL algorithm.

\subsection{Experiments for acquiring diversity}
In this subsection, we focus on quantitatively evaluating how much diversity the proposed method can provide to a single agent.
For this purpose, the behavior of peripheral agents other than the targeted agent is based on a fixed policy prepared by pre-training.
Besides, the peripheral agents run in a mode where they cannot observe the evaluated target agent not to interact with it.
Thus, the trajectories of other vehicles are fixed for each scenario.

\begin{table}
 \vspace{1mm}
 \centering
 \caption{Comparison of diversity scores for 50 policies. $|\Pi_{\geq 90\%}|$ is the number of candidates policies with a success rate 90\% or higher. Suc., O.A., and I.P. are the average success rate, overall diversity, and inter-policy diversity for selected 50 policies, respectively.
 Since {\em PolicySelect+DDE300} had only 25 filtered candidates, we selected all 25 policies instead.}
 \begin{tabular}{lr|rrrr} \hline
      Method                         & $|\Pi_{\geq 90\%}|$ & Suc.   & O.A. & I.P. \\ 
                                     &     & (\%) & (ave.) & (ave.) \\ \hline\hline
      {\em RandomTrajectories}       & N/A & N/A       & 1.49     & 3.44          \\\hline
      {\em RandomSelect300}          & 79 & 93.20     & 3.31    & 1.32         \\
      {\em PolicySelect300}          & 79 & 93.40     & 3.12    & 1.66         \\
      {\em PolicySelect+DDE300}   & 25 & 93.20     & 2.84    & 2.11         \\\hline
      {\em RandomSelect1200}         & 338  & 93.00     & 3.22    & 1.46         \\
      {\em PolicySelect1200}         & 338 & 94.32     & 2.72    & 2.24         \\
      {\em PolicySelect+DDE1200}  & 111 & 93.20     & 2.64    & 2.50         \\\hline
      {\em RandomSelect14400}        & 3926  & 93.56     & 3.25    & 1.48         \\
      {\em PolicySelect14400}        & 3926  & 92.88     & \textbf{2.39} & 2.90 \\
      {\em PolicySelect+DDE14400} & 1236 & 93.23     & 2.46    & \textbf{3.18}          \\ \hline
 \end{tabular}
 \label{tbl-compare-divesity-score-for-50}
 \vspace{-1mm}
\end{table}

\begin{figure*}
  \vspace{2mm}
  \begin{minipage}[b]{0.24\linewidth}
  \centering
  \includegraphics[keepaspectratio, width=4.0cm]
  {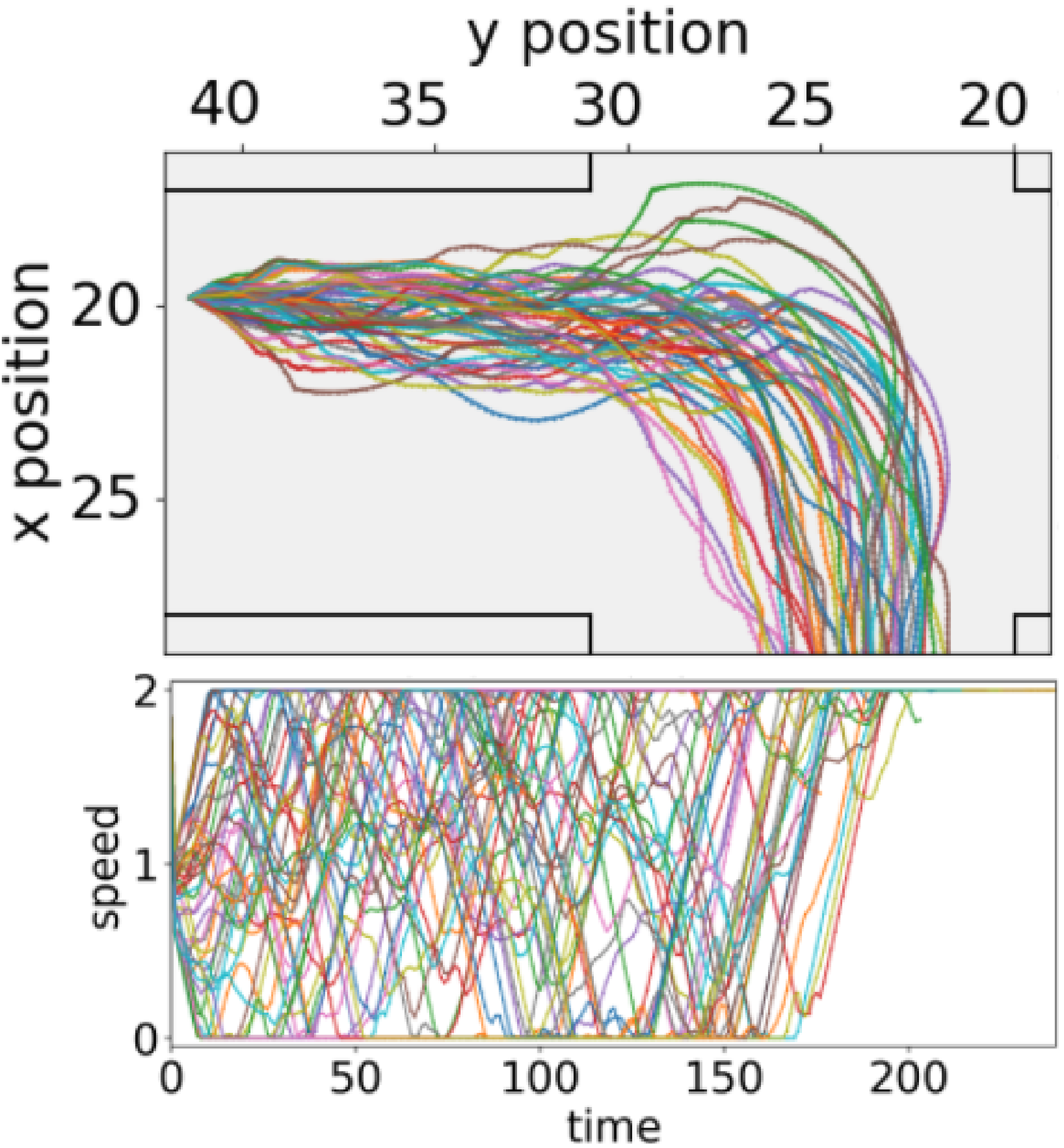}
  \\ {\small (O.A.: 1.49, I.P.: 3.44)}
  \subcaption{RandomTrajectory}\label{fig:reference_trajectory}
 \end{minipage}
 \hspace{1mm}
 \vline
 \hspace{1mm}
 \begin{minipage}[b]{0.24\linewidth}
  \centering
  \includegraphics[keepaspectratio, width=4.0cm]
  {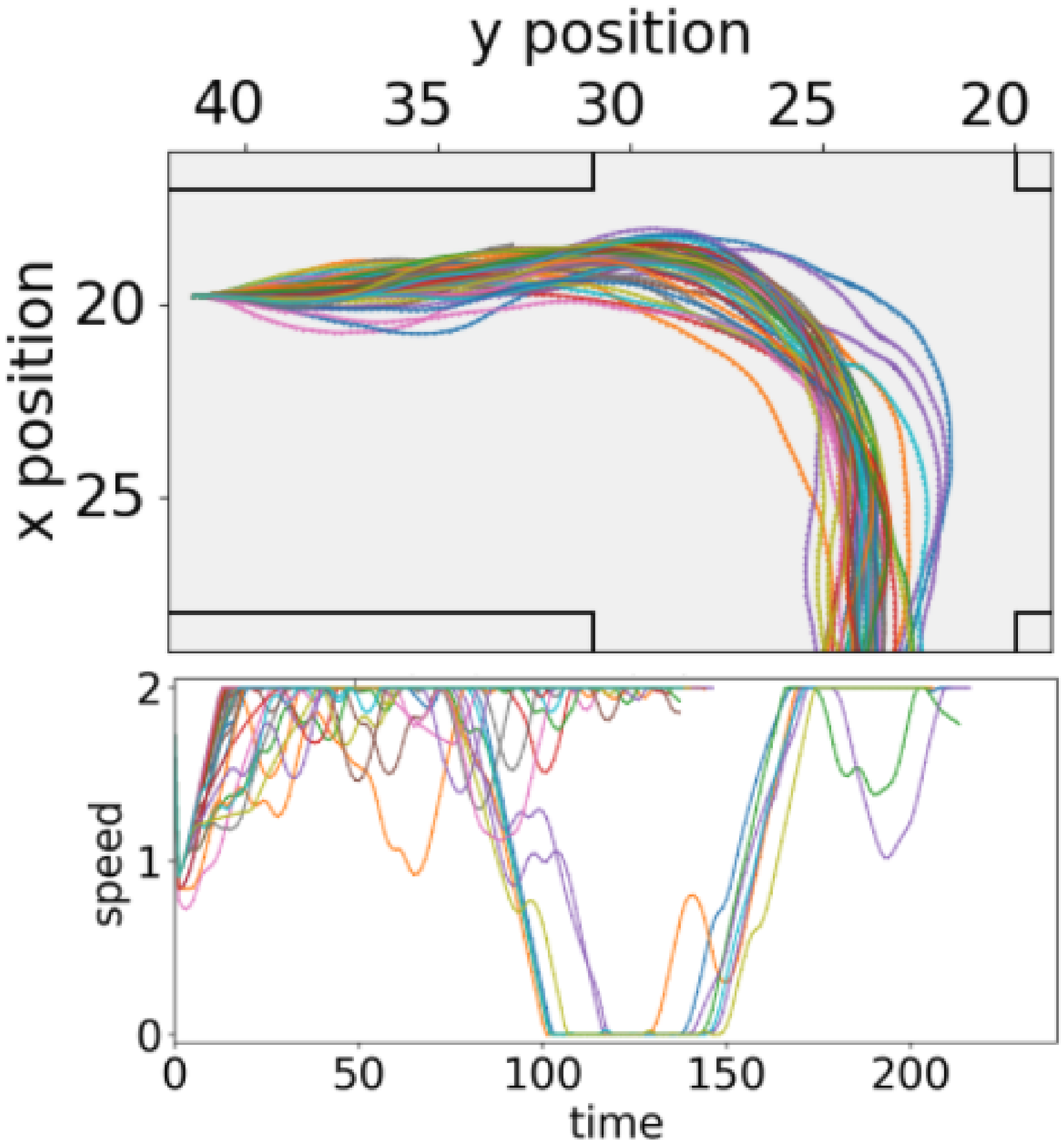}
  \\ {\small (O.A.: 3.22, I.P.: 1.32)}
  \subcaption{RandomSelect300}
 \end{minipage}
 \begin{minipage}[b]{0.24\linewidth}
  \centering
  \includegraphics[keepaspectratio, width=4.0cm]
  {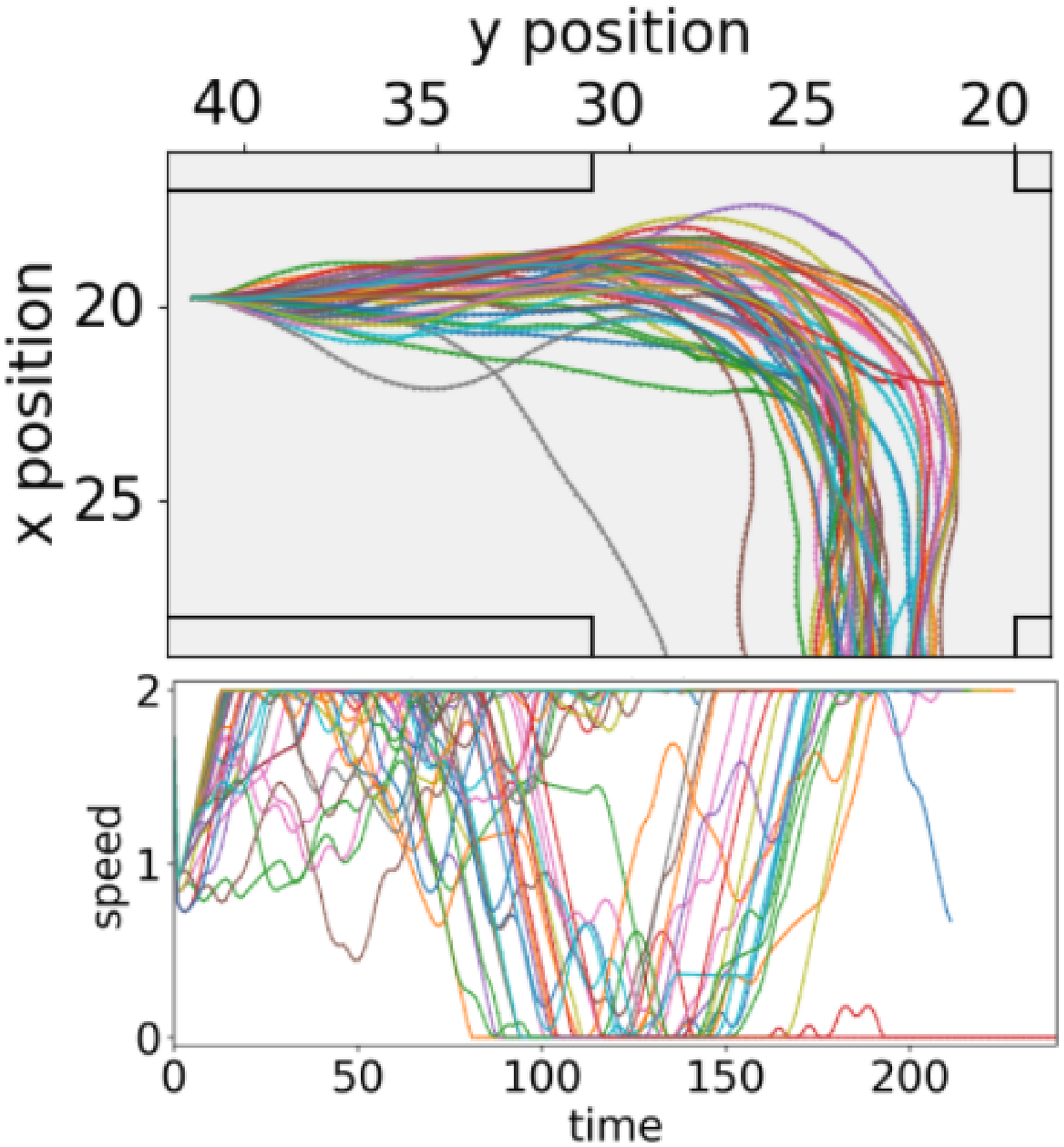}
  \\ {\small (O.A.: 2.72, I.P.: 2.24)}
  \subcaption{PolicySelect1200}\label{policyselect1200}
 \end{minipage}
 \begin{minipage}[b]{0.24\linewidth}
  \centering
  \includegraphics[keepaspectratio, width=4.0cm]
  {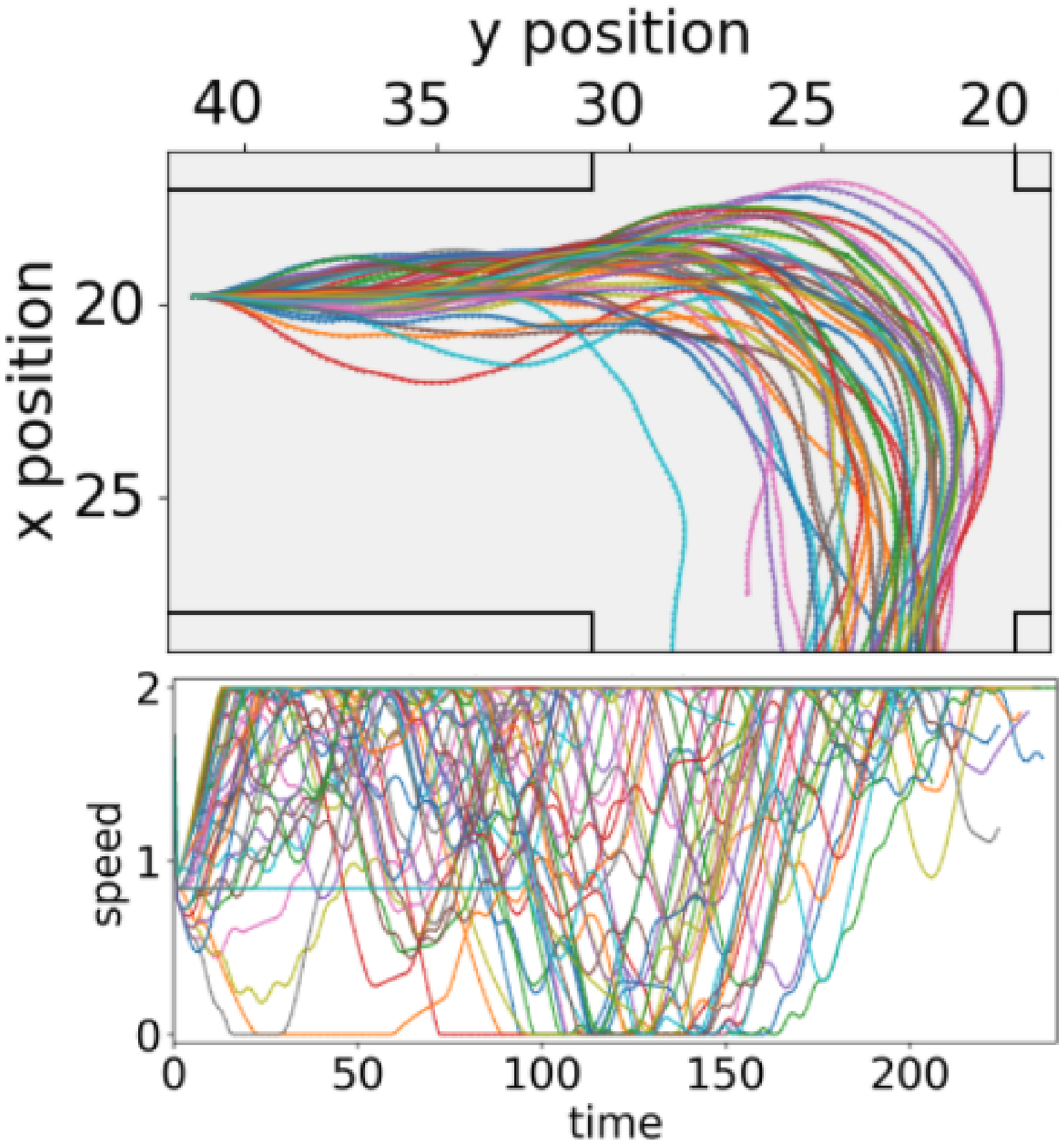}
  \\ {\small (O.A.: 2.46, I.P.: 3.18)}
  \subcaption{PolicySelect+DDE14400}
 \end{minipage}
 \caption{
 Visualizations of trajectories over 50 policies for a right-turn. The top represents positional changes (rotated clockwise by 90 degrees), and the bottom represents speed changes.}\label{vis-50policies}
 \vspace{-2mm}
\end{figure*}

The evaluation results are shown in Table~\ref{tbl-compare-divesity-score-for-50}. As previously described, larger inter-policy diversity means better, while smaller overall diversity means better.
As a reference, we added {\em RandomTrajectory}, a set of 50 trajectories generated with the Brownian bridge described in Section~\ref{subsec: overall_diversity} for each scenario.
Here, the parameter $\nu$ required for P-control is set to 2.0 [s].
The names beginning with \emph{RandomSelect}, \emph{PolicySelect}, and \emph{PolicySelect+DDE} are the results of extracted 50 policies by using the uniformly at random selector, our selector, and our selector whose candidates are trained with our modified DDE, respectively.
Each number appended to each method name represents the number of candidate policies.
Table~\ref{tbl-compare-divesity-score-for-50} shows that, as the number of candidate policies increases, \emph{PolicySelect} and \emph{PolicySelect+DDE} get better scores than \emph{RandomSelect} in both diversities.
Although \emph{PolicySelect14400} gets the best overall diversity, incorporating DDE gives a better score for the inter-policy diversity and competitive scores for the overall diversity whereas $|\Pi_{\geq 90\%}|$ is smaller than ones without DDE.
One possible reason is that the repulsive force induced by the intrinsic rewards helps to search spatially distinct trajectories, which leads to higher inter-policy diversity.

\begin{figure}
  \centering
  \includegraphics[clip,width=8cm]{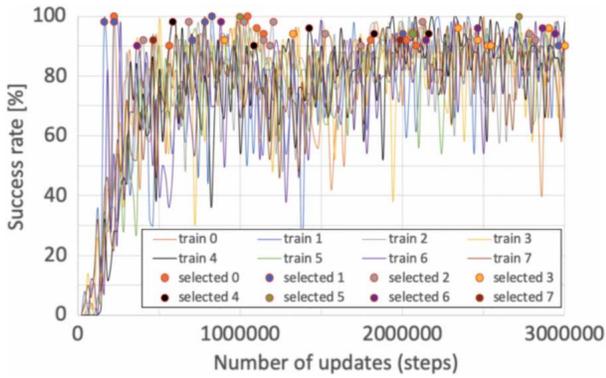}
  \caption{
    Changes in the success rate of 8 training sessions for \emph{PolicySelect1200}. The circle markers correspond to 50 policies selected in diverse policy selection. 
    Differences in marker colors indicate that policies were generated from different training sessions.
  }\label{fig:policy-select-vis}
\end{figure}

To show that our diversity measures correctly evaluate diversity, we also visualized the trajectories generated from trained 50 policies for a scenario in Fig.~\ref{vis-50policies}.
There seems to be a positive correlation between better diversities and trajectory coverage.
Especially, this shows that there is a great difference in the variety of speed changes.

We also visualized an example (\emph{PolicySelect1200}) to check the distribution of selected 50 policies generated from multiple training sessions.
As shown in Fig.~\ref{fig:policy-select-vis}, the policy selector employed a variety of training processes.
The success rate reached around 80\% within 200,000 steps then kept fluctuating strongly until the training completed.
One hypothesis is that the space of successful trajectories is multi-modal and the RL trainer finds various failures during the exploration.
Therefore, the policies in different training steps have different strategies, and actually Fig.~\ref{policyselect1200} indicates diverse behaviors.
That would be one possible reason why using past snapshots helps diversity acquisition.

\subsection{Experiments with multi-agent settings}

\begin{figure*}
  \vspace{2mm}
  \centering
  \includegraphics[keepaspectratio, width=18cm]{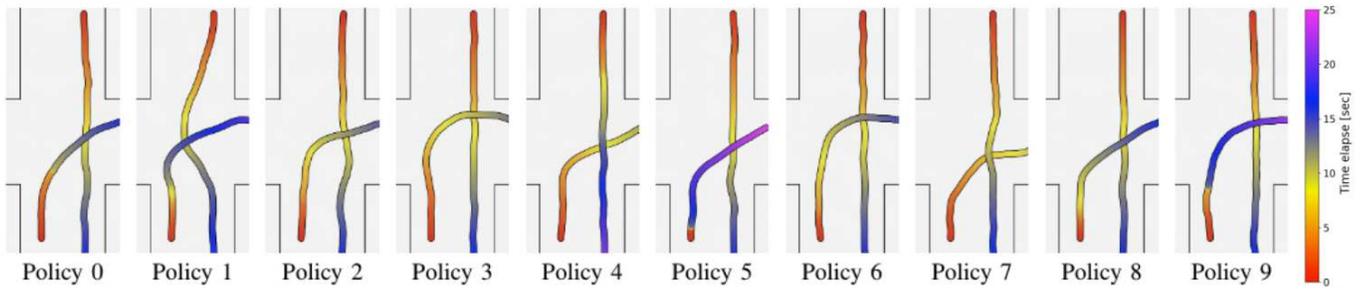}
  \caption{
  Behavioral differences for each policy in the same scenario.
  The color transition means the time change.
  Two points of the same color show the positions of two vehicles at the same time.
  }\label{fig:vis-one-scenario}
  \vspace{-2mm}
\end{figure*}

One last experiment we conducted with the purpose of checking if our RL-based policies can express a variety of interactive behaviors.
In this experiment, each policy learns behaviors of both the right-turn and oncoming vehicles.
Each vehicle is controlled in a decentralized manner by using a copy of the same policy.
In training, we used transitions of one vehicle whose index is switched for each episode.
We selected ten policies from 24,000 candidates by using our policy selector.

We examined the behavioral differences in the same scenario. As shown in Fig. \ref{fig:vis-one-scenario}, the selected policies display a variety of interactions:

\begin{itemize}
    \item Policies 0, 1, 2, 5, 8, 9 seem to yield the other vehicle.
    The yielding position differs in each case.
    Policies 5, 9 had long stops, while Policy 2 decelerated gradually.
    \item Policies 3, 6 gave way to the upcoming vehicle by taking a large detour to the left side.
    \item Policies 4, 7 took no yielding behavior. Especially, Policy 4 made the upcoming vehicle wait for the right-turn vehicle.
\end{itemize}

\section{Conclusion} \label{sec: conclusion}
We presented a method for acquiring diverse driving policies that leverages RL's exploration abilities and intrinsic rewards. Our experiments showed how we were able to acquire driving policies that optimize their behavioral diversity while maintaining good driving skills. 
Our proposed trajectory-based metrics to evaluate diversity, are also shown to be very helpful in building an efficient policy selection agnostic to how candidate policies are generated. We do expect that our approach could generate even more diverse policies by combining various policy generation methods. 
Such diverse driving modeling could also open new possibilities in building stronger in-car behavior prediction modules in the future.
Although this work utilized driving skills as the filter, it would be possible to generate human-like behaviors by comparing them with real-world driving data.
Since the parameters of reference trajectories are dependent on each simulation environment, an automatic parameter selector could help the diversity evaluation for other traffic scenes.

\section*{Acknowledgment}
This work was supported by Toyota Research Institute - Advanced Development, Inc.
We would like to thank Shin-ichi Maeda, Masanori Koyama, Mitsuru Kusumoto, Yi Ouyang, Daisuke Okanohara and anonymous reviewers for their helpful feedback and suggestions.
\bibliography{references}
\bibliographystyle{IEEEtran}

\end{document}